\documentclass[sigconf]{acmart}

\AtBeginDocument{%
  \providecommand\BibTeX{{%
    \normalfont B\kern-0.5em{\scshape i\kern-0.25em b}\kern-0.8em\TeX}}}

\renewcommand\footnotetextcopyrightpermission[1]{} 

\usepackage{algorithm}
\usepackage{algorithmic}
\usepackage{booktabs}
\usepackage{multirow}
\usepackage{tikz}
\usepackage{color}

\begin{document}
\fancyhead{}

\title{Generate Point Clouds with Multiscale Details from Graph-Represented Structures}


\author{Ximing Yang}
\affiliation{%
  \institution{School of Computer Science, Fudan University}
  \city{Shanghai}
  \country{China}}
\email{xmyang19@fudan.edu.cn}

\author{Zhibo Zhang}
\affiliation{%
  \institution{School of Computer Science, Fudan University}
  \city{Shanghai}
  \country{China}}
\email{zhibozhang21@m.fudan.edu.cn}

\author{Zhengfu He}
\affiliation{%
  \institution{School of Computer Science, Fudan University}
  \city{Shanghai}
  \country{China}}
\email{zfhe19@fudan.edu.cn}

\author{Cheng Jin}
\affiliation{%
  \institution{School of Computer Science, Fudan University}
  \city{Shanghai}
  \country{China}}
\email{jc@fudan.edu.cn}

\begin{abstract}
As details are missing in most representations of structures, the lack of controllability to more information is one of the major weaknesses in structure-based controllable point cloud generation. It is observable that definitions of details and structures are subjective. Details can be treated as structures on small scales. To represent structures in different scales at the same time, we present a graph-based representation of structures called the Multiscale Structure Graph (MSG). Given structures in multiple scales, similar patterns of local structures can be found at different scales, positions, and angles. The knowledge learned from a regional structure pattern shall be transferred to other similar patterns. An encoding and generation mechanism, namely the Multiscale Structure-based Point Cloud Generator (MSPCG) is proposed, which can simultaneously learn point cloud generation from local patterns with miscellaneous spatial properties. The proposed method supports multiscale editions on point clouds by editing the MSG. By generating point clouds from local structures and learning simultaneously in multiple scales, our MSPCG has better generalization ability and scalability. Trained on the ShapeNet, our MSPCG can generate point clouds from a given structure for unseen categories and indoor scenes. The experimental results show that our method significantly outperforms baseline methods.
\end{abstract}






\maketitle

\section{Introduction}
\label{sec:intro}

In real-world point cloud generation applications, the controllability to generate shapes is a fundamental requirement. Structures are the ideal intermediate representation to control the generation as they are the brief abstraction of 3D objects and have the advantages of conciseness and intuitiveness. Therefore, structure-based controllable point cloud generation is attracting rising research interest.

Previous works propose different structure representations—some work model objects at the part level. Mitra et al. \cite{Mitra2014} proposed structured objects containing part-level connectivity and inter-part relationship information. The information can be naturally formed as a part tree \cite{Mo_2019_CVPR} named N-ary part hierarchies. Mo et al. \cite{mo2019structurenet} proposed the StructureNet to generate point clouds via hierarchical graph networks, and later they proposed PT2PC \cite{mo2020pt2pc} to generate shapes directly from the part tree. Unlike the above methods, Yang et al. \cite{Yang_Wu_Zhang_Jin_2021} used a simple sparse point cloud to represent structures and introduced point-level semantic labels into structure extraction. They introduced CPCGAN to generate 3D shapes from a structure point cloud by extrapolating 64 points from each structure point.

Those works \cite{Mitra2014, Mo_2019_CVPR, mo2019structurenet, mo2020pt2pc, Yang_Wu_Zhang_Jin_2021} have achieved great performances in controllable point cloud generation from structures. But there are two major limitations in their works:
\begin{enumerate}
    \item \textbf{Lack of control in details.} Part-tree-based methods typically struggle to control generation within individual parts, while CPCGAN's representation of structures using only 32 points can limit the ability to capture small-scale details. The controllability emphasized in the methods mentioned earlier primarily applies to large-scale structures, with neural networks responsible for learning and generating finer details that are inherently less controllable.
    \item \textbf{Lack of generalization ability.} Some kinds of category-specific semantic information have been introduced into generation by the above works, which limits the ability to generate in more categories. Besides, absolute positions of points are used in their generation processes, causing those methods to be unable to work well when structures are rotated, scaled, or translated.
\end{enumerate}

\begin{figure*}[htb]
    \centering
    \includegraphics[width=\textwidth]{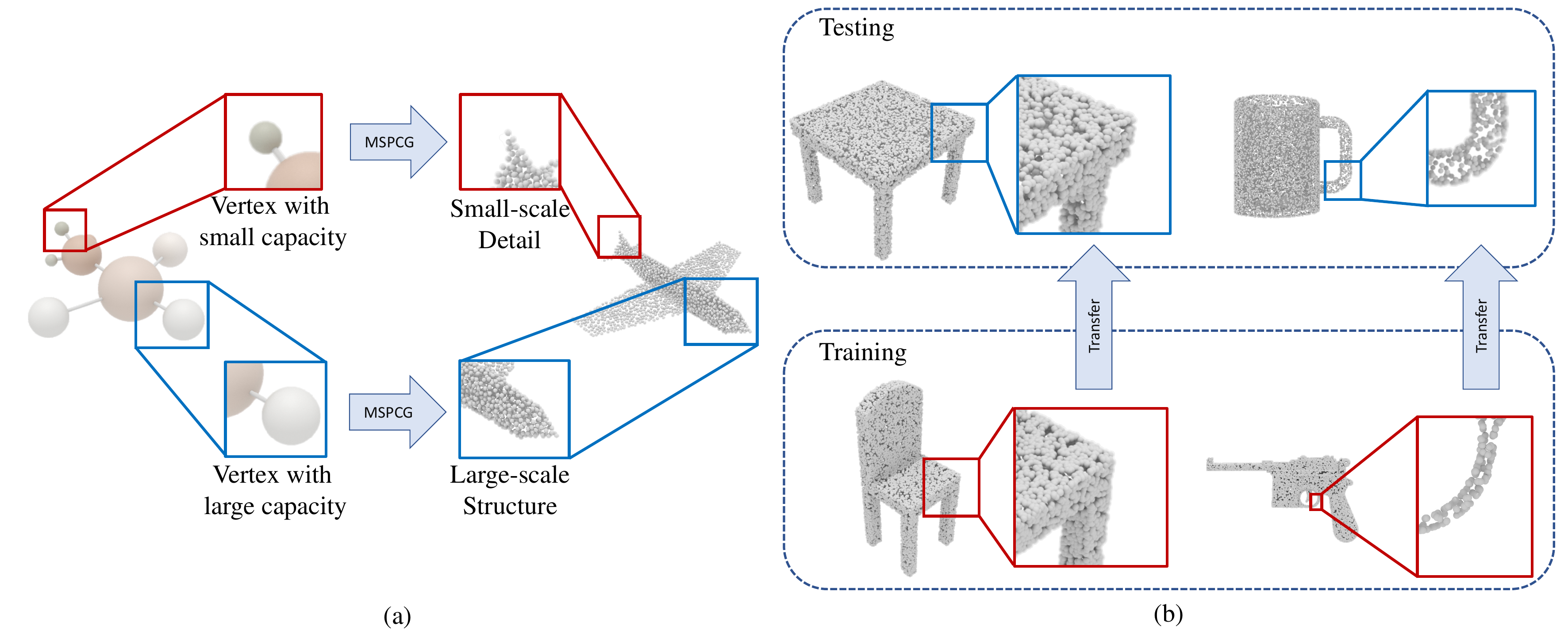}
    \caption{(a) An example of a Multiscale Structure Graph. Vertices with larger capacity imply more points in the generated point cloud and vice versa. (b) Examples of similar local structure patterns can help generate similar patterns in other categories and scales.} 
    \label{fig:ExampleofPatterns}
\end{figure*}

We propose a novel representation of structures aimed at addressing the challenge of controlling details. Our approach challenges the subjective distinction between details and structures, suggesting that details differ from structures mainly on a spatial scale. Given this assumption, details can be modeled similarly to structures. Furthermore, different parts of an object necessitate structure representations at varying scales; for instance, intricate parts benefit from relatively dense representations, while rough parts suffice with sparse representations. This calls for an ideal structure representation that is spatially uneven. However, most existing methods employ spatially uniform modeling of structures and struggle to handle spatially uneven structures. Therefore, we introduce a graph-based representation capable of modeling structures at different scales within a single graph, termed the Multiscale Structure Graph (MSG). As depicted in Fig.~\ref{fig:ExampleofPatterns}(a), the MSG represents structures continuously across scales by connecting vertices with different capacities corresponding to the numbers of points they represent. Fewer vertices with larger capacities suffice for modeling large-scale structures, while smaller capacities and more vertices better capture local shape details for small-scale structures.

Without considering physical or chemical influences, all 3D objects and parts can be scaled to any size without losing their rationality. Complex real-world 3D objects are often assembled by multiple simple shapes in different scales. Thus, as shown in Fig.~\ref{fig:ExampleofPatterns}(b), plenty of similar patterns in local structures can be found in 3D objects at different positions, scales, angles, densities, and categories. The knowledge learned from a pattern can be transferred to other similar patterns no matter the differences in those spatial properties. Previous methods cannot associate similar patterns as they use absolute positions of points that are sensitive to scaling and rotation. In this paper, we propose a novel Multiscale Structure-based Point Cloud Generator (MSPCG) to generate point clouds from local structures in a similarity-transformation-invariant (invariant to scale, rotation, and translation) manner. The invariances make MSPCG jointly learn similar patterns in different scales, which largely extends the scope of training data. As local structures are category-irrelevant and even dataset-irrelevant, learning the generation from local structures also helps the proposed method have better generalization ability.



The main contributions of our work are as follows:
\begin{enumerate}
    \item A concise and intuitive graph representation of structure called Multiscale Structure Graph (MSG) aiming to model a 3D object in different scales simultaneously. It enables the controlling of details during generation.
    \item A graph encoding and point cloud generation method called Multiscale Structure-based Point Cloud Generator (MSPCG). It generates point clouds from local structures and strongly improves the generalization ability towards scaling and rotation.
    \item A framework to simultaneously learn from different scales, positions, and angles. It extends the scope of training data and helps the MSPCG to perform competitively in point cloud reconstructions compared to SOTA methods with complete point clouds as input.
    \item The combination of the above contributions has great generalization capabilities crossing categories and datasets. Trained on ShapeNet, our method can generate point clouds from given structures for unseen categories, objects from ModelNet, and indoor scenes from ScanNet.
\end{enumerate}

\begin{figure*}[htb]
    \centering
    \includegraphics[width=\textwidth]{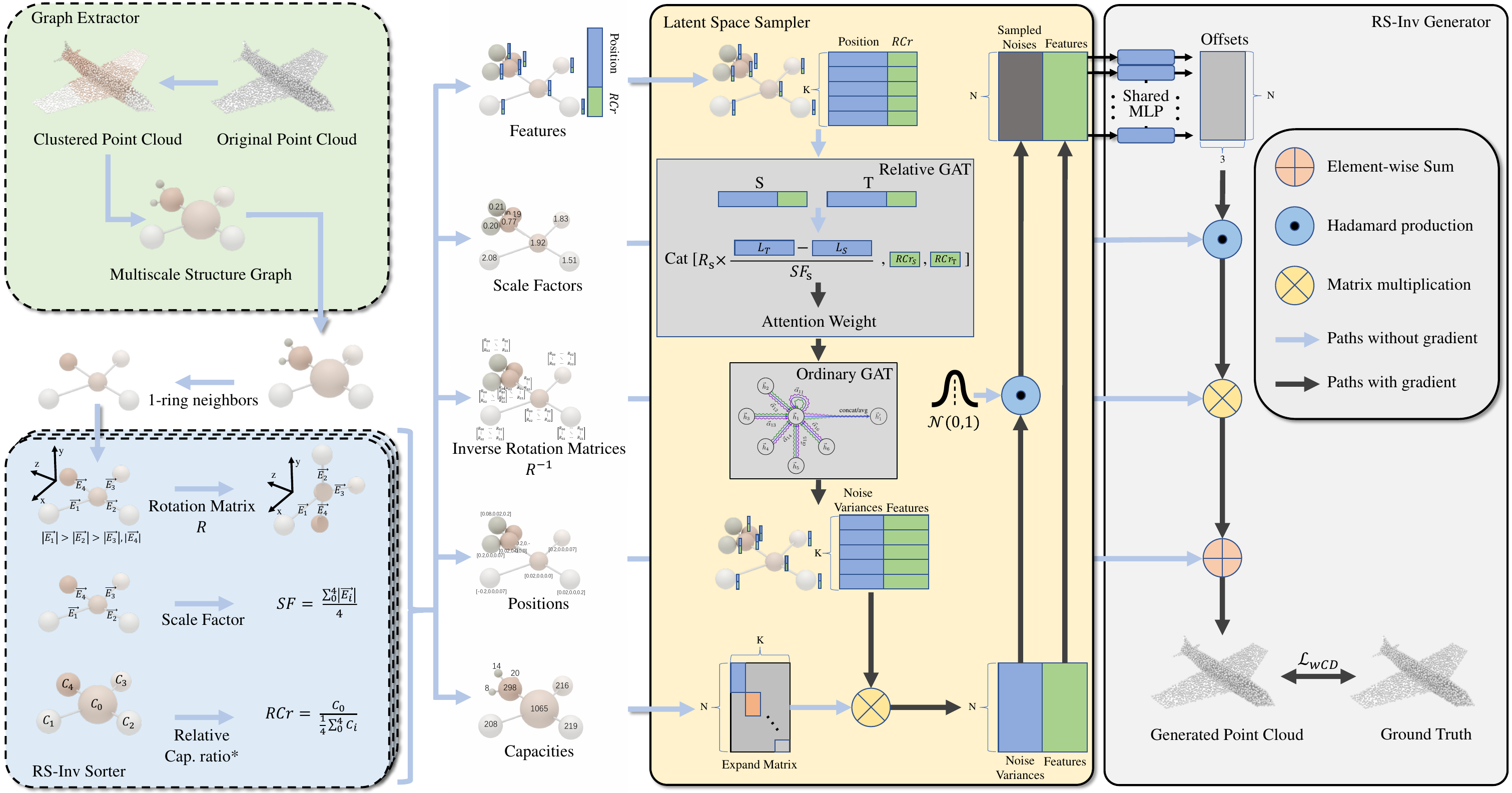}
    \caption{Overview framework of proposed MSPCG. K is the number of vertices and N is the number of points we finally generate. $ ^* $Relative Capacity ratios are calculated in 3-ring neighbors because the depth of GAT is 3.}
    \label{fig:Network}
\end{figure*}
    
\section{Related works}
\label{sec:related works}
    
\subsection{Random Point Cloud Generation}
After various researches have been done in 1D signal and 2D image\cite{goodfellow2020generative, kingma2018glow, kingma2013auto, oord2016conditional, prenger2019waveglow, shen2018natural}, methods of using a neural network to generate 3D point clouds have been explored in recent years. Achlioptas et al. \cite{achlioptas2017learning} firstly adopted simple fully connected layers as the generator to encode and generate point clouds. Valsesia et al. \cite{valsesia2018learning} used dynamic graph convolution networks to enhance the generation performance. Shu et al. \cite{shu20193d} proposed TreeGAN, which generates point clouds from a tree-based network. Yang et al. \cite{yang2018foldingnet} proposed FoldingNet deforming 2D squares into 3D surfaces to generate point clouds. AtlasNet\cite{groueix2018} followed the idea of FoldingNet and further expanded the deforming operation into multiple branches. TearingNet\cite{pang2021tearingnet} researched modeling shapes with more complex topology. PointFlow \cite{pointflow} modeled the point cloud generation as a distribution transformation by introducing free-form normalizing flows\cite{chen2018neural, grathwohl2019ffjord}.

The above methods aim to generate point clouds from random latent codes. 

\subsection{Controllable Point Cloud Generation}
To enhance control and detail in the generation, certain researchers have concentrated on producing point clouds using various types of intermediate representations. Some works \cite{Mitra2014, mo2020pt2pc, Mo_2019_CVPR} controllably generate point clouds from part-level structure representations. Mitra et al. \cite{Mitra2014} disassembled a 3D object into several connected parts, constructing a part-level structure representation. Mo et al. \cite{Mo_2019_CVPR, mo2020pt2pc} proposed StructureNet and PT2PC methods to controllably generate point clouds from the part-level structure representations. However, the part-level structure representation is expensive to annotate. Yang et al. \cite{Yang_Wu_Zhang_Jin_2021} constructed structure point clouds using point-level semantic annotations. The CPCGAN they introduced is also controllable and capable of generating semantic labels for points.

A popular way to explicitly construct rotation-insensitive neural networks is using capsule networks\cite{sabour2017dynamic}. Researchers adopted capsule networks in 3D point cloud reconstruction task\cite{sun2021canonical, zhao20193d}. Unlike capsule networks, which force the network to estimate the poses while learning embeddings of objects, our method explicitly extracts pose information. Apart from the invariances of rotation, our method is also able to generate reasonable shapes when inputs are scaled.


\section{Approach}
\label{sec:methods}
    
\subsection{Overview}
The overview of our MSPCG is shown in Fig.~\ref{fig:Network}. The Multiscale Structure Graph (MSG) is presented as $ G_{MS}(V, E) $, where each vertex $V_i$ contains a $C_i$ standing for the capacity and an $L_i$ meaning the location coordinates. Given a $ G_{MS} $ with $K$ vertices, we aim to generate a dense 3D point cloud $P_{ge} = \{g_j\}_{j=1}^{N}$ where $g_j \in \mathbb{R}^3$ and $N =  \sum_{i=1}^{K} C_i$ is the number of points we want to generate. In training, each $ G_{MS} $ has a corresponding ground truth point cloud $P_{gt} = \{gt_j\}_{j=1}^{N}$. The goal of MSPCG is to generate $P_{ge}$ closer to $P_{gt}$.
There are mainly four main parts in the MSPCG:
\begin{enumerate}
    \item The graph extractor samples a $G_{MS}$ from the $P_{gt}$. 
    \item The rotation and scale invariant sorter (RS-Inv sorter) takes a $G_{MS}$ as input and then extracts multiple properties on each vertex.
    \item The latent space sampler first encodes the $G_{MS}$ in a similarity-transformation-invariant manner via a modified graph attention network. Then an expanding operation is performed to sample $N$ features where each feature represents one point in the final point cloud.
    \item The RS-Inv generator projects the features from latent space into 3D space, getting the $P_{ge}$ finally by generating a bunch of points around each vertex in MSG. Besides, to balance the influences caused by the different capacities of vertices, a weighted Chamfer Distance is proposed as the loss function of our MSPCG.
\end{enumerate}
    
\subsection{Multiscale Structure Graph}
To model details of 3D objects in different scales, we propose a graph-based structure representation $ G_{MS} $ as shown in Fig.~\ref{fig:ExampleofPatterns}(a). Each vertex $V_i$ is an abstraction of a point cluster in the original point cloud. $L_i \in \mathbb{R}^3$ represents the center of the cluster and the capacity $ C_i \in \mathbb{N}^+$ is the size of the cluster. Edges $E$ between vertices represent the connectivities of the clusters in 3D space. By this definition, we consider large-scale structures as inadequately described details and regard low-level details as excessively described structures.
    
\subsection{Graph Extractor}
To enhance the generalization ability, graphs containing multiscale structures are required in training. Yang et al. \cite{Yang_Wu_Zhang_Jin_2021} utilize the K-Means algorithm to get point clusters and then extract a structure point from every cluster. Following this idea, we propose a simple and efficient algorithm called Mixed-Precision Random K-Means to extract point clusters within different sizes. Pseudo-code of the algorithm is given in Algorithm ~\ref{alg1}. The gravity center of cluster $i$ will be the $L_i$ in $G_{MS}$, and the $C_i$ is assigned by the number of points in the cluster $i$. Vertices are connected by an edge if the spatial distance between the clusters they represent is closer than a threshold. With the mixed-precision K-means, we can automatically extract MSGs that contain structures in various scales. The randomness in the algorithm also improves the robustness of MSPCG to different graph distributions.

\begin{algorithm}
    \caption{Mixed-Precision Random K-Means}
    \label{alg1}
    \begin{algorithmic}
        \STATE \textbf{Input:} $P_{gt} = \{gt_j\}_{j=1}^{N}\ where\ gt_j \in \mathbb{R}^3$
        \STATE $N_c \gets RandInt(4,16),\ N_f \gets RandInt(64,128)$
        \STATE $\mathit{Centroids}_{coarse} \gets KMeans(P_{gt}, N_c),\ \mathit{Centroids}_{fine} \gets KMeans(P_{gt}, N_f)$
        \STATE $N_{f2c} \gets RandInt(12,32)$
        \STATE $\mathit{Centroids}_{f2c} \gets RandChoice(\mathit{Centroids}_{fine}, N_{f2c})$
        \STATE $\mathit{Clusters} \gets GetClustersFromCentroids(P_{gt}, \mathit{Centroids}_{f2c})$
        \STATE \textbf{Output:} $\mathit{Clusters}$
    \end{algorithmic}
\end{algorithm}


\subsection{RS-Inv Sorter}
To encode the graph in a similarity-transformation-invariant manner, a rotation matrix $R_i$ and a scale factor $\mathit{SF}_i$ are needed for every vertex $V_i$ in the graph. As the capacity of each vertex is also scale-related, directly feeding capacities into the network will also damage the scale invariances of the model. Thus we extract another property called relative capacity ratio $\mathit{RCr}_i$ for each vertex $ V_i$ in the RS-Inv sorter. Assume $EC_i = \{E_j\},\ E_j = (V_i, V_n), V_n \in \mathcal{N}_i$ is the set of edges connected to $V_i$. We define the edge vector $\overrightarrow {E_j} = L_n - L_i$ for $E_j$ where $\overrightarrow {E_j} \in \mathbb{R}^3$. Without loss of generality, we suppose the two longest non-collinear edge vectors in $EC_i$ are $E_1$ and $E_2$. $R_i$ is a rotation matrix that meets the following conditions:
\begin{equation}
    \begin{aligned}
        &R_{i} \times  \frac{\overrightarrow {E_1}}{|\overrightarrow {E_1}|} = \begin{bmatrix} 1 & 0 & 0 \end{bmatrix}^T\ &if\ |EC_i| > 0 \\
        and\ &R_{i} \times  \overrightarrow {E_2} \times \begin{bmatrix} 0 & 0 & 1 \end{bmatrix} =  0 \ &if\ |EC_i| > 1 \\
        and\ &R_{i} \times  \overrightarrow {E_2} \times \begin{bmatrix} 0 & 1 & 0 \end{bmatrix} > 0 \ &if\ |EC_i| > 1
    \end{aligned}
\end{equation}
If $|EC_i| = 0$, the $R_i$ is an identity matrix.

The scale factor for $V_i$ is calculated as:
\begin{equation}
    \mathit{SF}_i = \frac{\sum_{E_j \in EC_i}|\overrightarrow {E_j}|}{|EC_i|} \ \ \ \ \ if\ |EC_i| > 0 
\end{equation}
if $|EC_i| = 0$, $\mathit{SF}_i$ is the average length of all edges.

The relative capacity ratio for $V_i$ comes from:
\begin{equation}
    \mathit{RCr}_i = \frac{C_i}{\frac{1}{|N(3)_i|}\sum_{V_j \in N(3)_i}C_j}
\end{equation}
Where $N(3)_i$ is the set of vertices whose graph distance to $V_i$ is smaller than 3.

\subsection{Latent Space Sampler}
With a $G_{MS}$ as input, the latent space sampler aims to encode the graph with $K$ vertices and sample $N$ features in latent space. Graph encoders have been widely researched in recent years\cite{perozzi2014deepwalk, bruna2013spectral, hamilton2017inductive, velivckovic2017graph}. The Graph Attention Network\cite{velivckovic2017graph}(GAT) is one of the most popular message-passing-based graph neural networks. GAT can efficiently aggregate features from neighbors and avoid the over-smooth problem. Thus we choose GAT as the backbone of our graph encoder.

To avoid the usage of absolute positions, we propose a relative graph attention layer (relative GAT) as the first layer of the graph encoder.
The attention weight $\alpha$ between $V_s$ and $V_t$ is expressed as:
\begin{equation}
    \alpha_{st} = \frac{e^{\mathit{LR}(\overrightarrow {a}^T[R_s\times \frac{L_t - L_s}{SF_s} \parallel RCr_s \parallel RCr_t])}}{\sum_{k \in \mathcal{N}_s}e^{\mathit{LR}(\overrightarrow {a}^T[R_s\times \frac{L_k - L_s}{SF_s} \parallel RCr_s \parallel RCr_k])}} 
\end{equation} 
where $\overrightarrow {a}\in \mathbb{R}^5$ is the weight vector, $\parallel$ is the concatenation operation, and $\mathit{LR}$ stands for LeakyReLU nonlinearity.
The feature vectors $F_{st}$ for $V_t$ that will be aggregated by the attention weights of $V_s$ is expressed as:
\begin{equation}
    F_{st} = [R_s\times \frac{L_t - L_s}{SF_s} \parallel RCr_s \parallel RCr_t]
\end{equation}
The introduction of the relative capacity ratio helps the network to recognize the differences between vertices representing different scales.
At the same time, the relative capacity ratio only considers vertices in a certain range, which is also robust to local density changes. After relative GAT, two layers of ordinary GAT with layer normalizations are added to further encode the graph.

The encoder outputs noise variances $\mathit{NV}_i \in \mathbb{R} ^{c}$ and encoded features $F_i \in \mathbb{R}^{c}$ for each vertex $V_i$, where $c$ stands for the number of feature channels. To generate a point cloud for $N = \sum_{i=1}^{K} C_i$ points, each vertex $V_i$ needs to breed $C_i$ points. Most multi-branch generation methods in previous works can only generate the same number of points in every branch\cite{groueix2018, mo2020pt2pc, Yang_Wu_Zhang_Jin_2021}. FoldingNet\cite{yang2018foldingnet} proposed a sample-transform method, which can generate a shape in different numbers of points. With the inspiration FoldingNet gave to us, we propose a method to sample different numbers of features for different vertices.
A binary expanding matrix $\mathit{EP} \in \mathbb{N}^{N*K}$ is presented:
\begin{equation}
    \mathit{EP}_{ij} = \left\{
        \begin{array}{rcl}
            1 & & i \ge \sum_{k = 0}^{j-1}C_k\ and\ i < \sum_{k=0}^j C_k\\
            0 & & otherwise\\
        \end{array}
    \right.
\end{equation}
By pre-multiplying $EP$ to the features $F \in \mathbb{R}^{K \times c}$, we can expand the $K$ features into $N$ features $F \in \mathbb{R}^{N \times c}$, where features from vertex $V_i$ will be repeated $C_i$ times. Pre-multiplying other properties matrices (e.g. $R,\mathit{SF}, L$) by $EP$ can also expand other properties in the same way. Similar to FoldingNet's sampling results, each feature is concatenated by a unique noise and a common global feature. Points expanded from $V_i$ share the global feature. Noises for these points are sampled by a reparameterization module from a Gaussian distribution whose variances are $\mathit{NV}_i$.

\begin{table*}[htb]
    \begin{tabular}{c|c|cccccc}
        \toprule[2pt]
        Methods                                                                             & \begin{tabular}[c]{@{}c@{}}Input\end{tabular}                                         & \begin{tabular}[c]{@{}c@{}}ShapeNet\\ Seen\end{tabular} & \begin{tabular}[c]{@{}c@{}}ShapeNet\\ Unseen\end{tabular} & \begin{tabular}[c]{@{}c@{}}ShapeNet\\ Unseen +\\ R-rotate\end{tabular} & \begin{tabular}[c]{@{}c@{}}ShapeNet\\ Unseen +\\ R-scale\end{tabular} & \begin{tabular}[c]{@{}c@{}}ShapeNet\\ Unseen +\\ R-rotate +\\R-scale\end{tabular}  & ModelNet       \\ \midrule[1pt]
        AtlasNet V2\cite{deprelle2019learning}                                              & \multirow{4}{*}{\begin{tabular}[c]{@{}c@{}}Point\\ Cloud\end{tabular}}                & 6.107                                                   & 6.988                                                     & 7.096                                                                  & 7.602                                                                 & 7.714                                                                              & 2049           \\ \cline{1-1} \cline{3-8} 
        TearingNet \cite{pang2021tearingnet}                                                &                                                                                       & 6.924                                                   & 6.767                                                     & 6.819                                                                  & 7.322                                                                 & 7.397                                                                              & 36.25          \\ \cline{1-1} \cline{3-8} 
        3D Point Capsules \cite{zhao20193d}                                                 &                                                                                       & 6.947                                                   & 7.626                                                     & 7.718                                                                  & 8.317                                                                 & 8.464                                                                              & 88.72          \\ \cline{1-1} \cline{3-8} 
        Canonical Capsules \cite{sun2021canonical}                                          &                                                                                       & \textbf{4.386}                                          & \textbf{5.307}                                            & \textbf{5.317}                                                         & \textbf{5.729}                                                        & \textbf{5.713}                                                                     & 46.50          \\ \midrule
        MSPCG w/o rotate\&scale                                                             & \multirow{2}{*}{MSG}                                                                  & 6.354                                                   & 7.281                                                     & 7.263                                                                  & 8.037                                                                 & 7.993                                                                              & 49.30          \\ \cline{1-1} \cline{3-8}
        MSPCG(ours)                                                                         &                                                                                       & 5.177                                                   & 6.323                                                     & 6.312                                                                  & 6.757                                                                 & 6.781                                                                              & \textbf{27.06} \\
        \bottomrule[2pt]
    \end{tabular}
    \caption{Average reconstruction performance in Chamfer Distance, multiplied by $10^4$. "R-" represents random.} 
    \label{table:CompareToSOTA}
\end{table*}
    
\subsection{RS-Inv Generator}
The RS-Inv generator firstly uses a shared MLP to get offsets $O \in \mathbb{R}^{N \times 3}$ by projecting features and sampled noise into 3D space.
The generated point cloud $P_{ge}  = \{g_j\}_{j=1}^{N}, g_j \in \mathbb{R}^{3}$ can be expressed as:
\begin{equation}
    g_j = R_j^{-1} \times O_j \cdot \mathit{SF}_j  + L_j 
\end{equation}                                                         

\subsection{Weighted Chamfer Distance Loss}
To learn in multiple scales simultaneously, each vertex should get the same attention no matter which scale it is located. But obviously, the vertices with bigger capacities will get more attention as the Chamfer Distance\cite{fan2017point} treats equally to every point rather than to every vertex. Thus, we propose the weighted Chamfer Distance $\mathcal{L}_{wCD}$ as the loss function to rebalance the influences of different capacities.
The weighted Chamfer Distance $\mathcal{L}_{wCD}$ between $P_{ge}  = \{g_i\}_{i=1}^{N}$ and $P_{gt} = \{gt_i\}_{i=1}^{N}$ is defined as:
\begin{equation}
    \begin{aligned}
        \mathcal{L}_{wCD}(P_{ge}, P_{gt}) &= \frac{1}{|P_{gt}|} \sum_{gt_j\in P_{gt}} \mathop{min}_{g_i \in P_{ge}}\parallel gt_j - g_i\parallel_2\\
        &+ \frac{1}{K} \sum_{g_i\in P_{ge}} \mathop{min}_{gt_j \in P_{gt}}(\frac{\parallel gt_j - g_i\parallel_2}{C_i})
    \end{aligned}
\end{equation}
As each point in $P_{ge}$ may not only be matched once, weighting them may cause the summation of weights not equal to 1. Therefore the first term in $\mathcal{L}_{wCD}$ is not balanced by the capacity. Thus weighted Chamfer Distance can only alleviate the problem but still cannot solve it completely.

\subsection{Invariance for similarity transformation and Learning from Multiple Scales}
Similarity transformation includes translation, rotation, and scaling. MSPCG gets the invariances for similarity transformation by:
\begin{enumerate}
    \item Giving the neural network a consistent input after a similarity transformation is adopted to the MSG.
    \item Applying inverse transformations to the output of the neural network to match the ground truth for calculating losses.
\end{enumerate}

From the perspective of one vertex in the relative GAT, the information from its neighbors is the rotated, normalized relative positions, and the relative capacity ratio. Vertices will get different information from the same neighboring vertex. RS-Inv sorter extracts a unique similarity transformation for each vertex, which will transform the vertex and its neighbors to a canonical pose. Therefore, the input of the attention module will be consistent no matter what similarity transformation is applied to the input MSG. As the transformation for a vertex is explicitly calculated, the RS-Inv generator directly applies an inverse transformation to the offsets generated from this vertex.

After the relative GAT wipes similarity transformation, other GATs will only focus on the topological and feature-wise relationship between vertices. Thus, information from different scales can be learned simultaneously by the shared network in our method.
    
\section{Experiments}
\label{sec:experiments}
    
\subsection{Dataset and Implementation Details}
We train our MSPCG on the ShapeNetCore-v2-PC15K\cite{shapenet2015, pointflow} dataset.     In training, we down-sample the 15,000 points into 2048 points using the farthest points sampling method\cite{qi2017pointnetplusplus} to be the $P_{gt}$.     Categories with more than 300 samples in the training split of ShapeNet are collected to be the training set. Thus some categories are unavailable for the model in the training period. We sample 5 MSGs for every point cloud to overcome the randomization of Mixed-Precision Random K-Means.

To prove the generalization of the proposed method, we introduce the ModelNet\cite{zhirong15CVPR} as an object-level dataset and the ScanNet\cite{dai2017scannet} as a scene-level dataset.

On implementation of the MSPCG, Adam optimizer\cite{kingma2014adam} is used with $\alpha = 1e-4,\ \beta_1 = 0$ and $\beta_2 = 0.99$. The model is trained for 90 epochs parallelly on 3 GPUs with batch size set to 32 on each card. The bottleneck for increasing batch size is the inference time. As each input data may have a different number of vertices, the time complexity of multiplying expanding matrix to features is $\mathcal{O} ((\sum K)\times(\sum N)\times C)$ where C is the number of feature channels. Increasing batch size will cause the inference time to squarely increase. As there is no relationship between different data samples, we split the big batch into different processes on different GPUs to better use the compute resources. Each process with a batch size of 32 will cost about 5 GB of GPU memory in training.

\subsection{Comparisons in Point Cloud Reconstruction}
As the MSPCG generates point clouds from given structures, it is unfair to compare MSPCG with point cloud generation methods that generate from random latent codes. MSPCG can be regarded as a reconstruction model that reconstructs complete point clouds from spare structures. Therefore, we compare our MSPCG mainly on the point cloud reconstruction task. Six testing sets of ShapeNet are constructed in experiments. The average Chamfer Distances between the generated point clouds and the ground truths are reported. All the testing sets come from the official testing split of ShapeNet. Without any other declaration, the input MSG in those testing sets is extracted by a K-Means algorithm with K randomly chosen from 16 to 64. The ShapeNet Seen includes the categories that are used in the training period, and the ShapeNet Unseen only contains the other categories. The suffix "32 vertices" means the K in the K-Means algorithm is restricted to 32. The R-rotate stands for randomly rotating each object and the R-scale stands for randomly scaling each object with a ratio between 0.8 and 1.25.

\begin{figure}[htb]
    \centering
    \includegraphics[width=\linewidth]{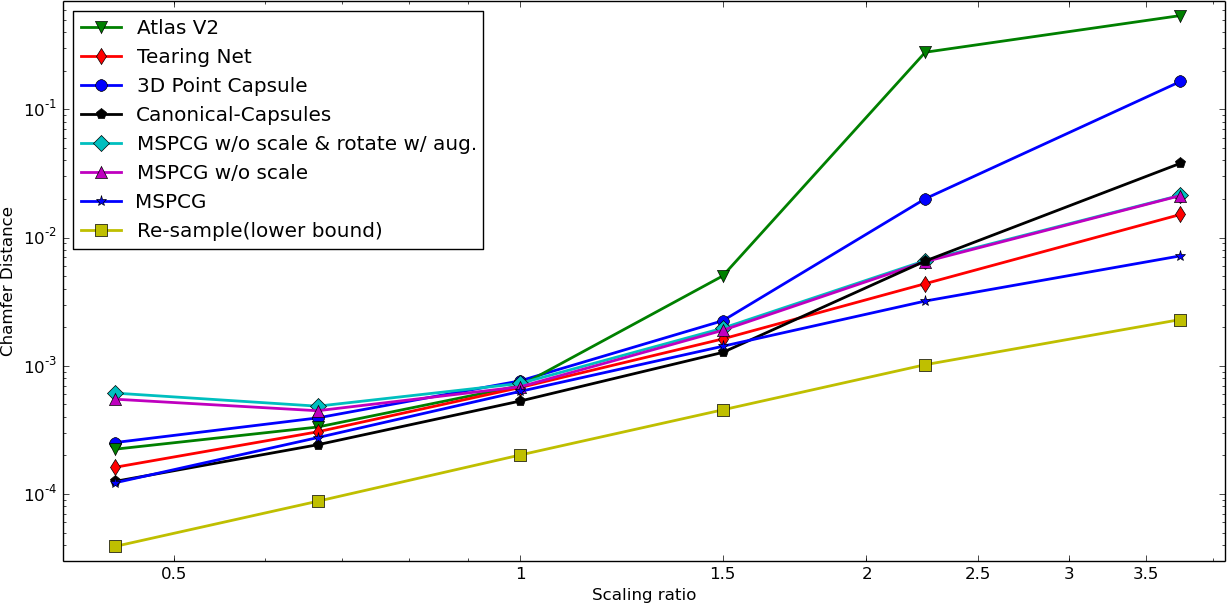}
    \caption{Reconstruction Chamfer Distances in ShapeNet Unseen when all point clouds are scaled in a certain ratio.}
    \label{fig:Scale-CD}
\end{figure}

We first compared our MSPCG to some state-of-the-art point cloud reconstruction methods in Tab.~\ref{table:CompareToSOTA}. Augmented data (randomly rotated and scaled) are used to train those methods. It can be found that although the input of MSPCG contains less information, our method can provide competitive or even better reconstruction results. We believe that it can be attributed to the strong representation ability of MSG and the simultaneous learning of local point patterns at multiple scales. When the scale changes more, it is more difficult to reconstruct a reasonable shape. Fig.~\ref{fig:Scale-CD} shows the reconstruction of Chamfer Distances when all point clouds are scaled at some specific ratios. We re-sampled 2048 points from the original 15,000 points for all point clouds and reported the Chamfer Distances between re-sampled point clouds and ground truth point clouds as the lower bound of the distance. As revealed by the figure, most methods other than MSPCG cannot provide consistent reconstruction performance when the inputs are zoomed. The Chamfer Distance growth trend of MSPCG is the same as the re-sampled point clouds, which illustrates the MSPCG is insensitive to scale changes. To verify the cross-generalization ability, we adopt those models trained on ShapeNet into the reconstruction task on ModelNet. The testing set of ModelNet-40 is used to compare the methods, and all objects are randomly rotated. The experimental results show that the proposed MSPCG has a much better cross-generalization ability than other methods. 

\subsection{Qualitative Comparison}

\begin{figure}[htb]
    \centering
    \includegraphics[width=\linewidth]{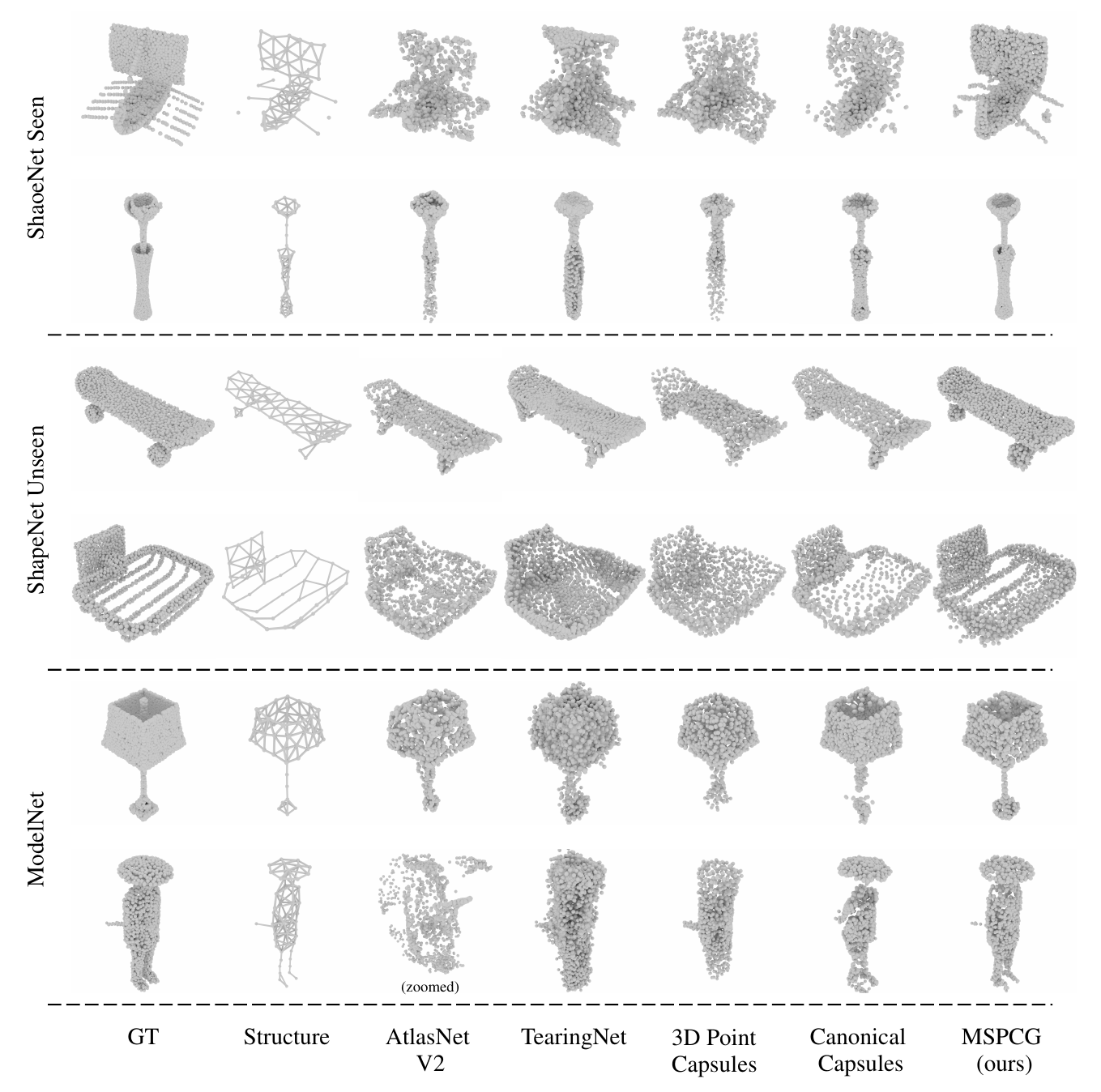}
    \caption{Reconstruction results on ShapeNet and ModelNet. Note that our MSPCG takes structure graphs as input and other methods use point clouds.}
    \label{fig:ShowResults}
\end{figure}

In addition to quantitative comparisons, we also conducted qualitative assessments by comparing our method with state-of-the-art (SOTA) methods. Some reconstruction results are showcased in Figure \ref{fig:ShowResults}. Other methods often prioritize reconstructing the large-scale shapes of the input and struggle to generate sharp details. This disparity becomes more pronounced when tested on different datasets.

MSPCG consistently produces visually impressive outputs that closely resemble the originals. This success is notable considering the significant differences in point coordinate distributions between ModelNet and ShapeNet. Neural networks without specialized designs for handling scale variations struggle with these distinct distributions due to their unfamiliarity. However, MSPCG overcomes this challenge by leveraging similarity-transformation-invariant designs and generating point clouds from local structures. This approach effectively aligns the new distributions encountered in ModelNet with those observed in ShapeNet, leading to the generation of plausible and accurate results.

\subsection{Ablation Studies}

\begin{table*}[htb]
    \begin{tabular}{@{}cccccccc@{}}
        \toprule[2pt]
                                                                                    & \begin{tabular}[c]{@{}c@{}}ShapeNet\\ Seen\end{tabular} & \begin{tabular}[c]{@{}c@{}}ShapeNet\\ Unseen\end{tabular} & \begin{tabular}[c]{@{}c@{}}ShapeNet\\ Unseen\\ 32 vertices\end{tabular} & \begin{tabular}[c]{@{}c@{}}ShapeNet\\ Unseen +\\ R-rotate\end{tabular}      & \begin{tabular}[c]{@{}c@{}}ShapeNet\\ Unseen + \\ R-scale\end{tabular}      & \begin{tabular}[c]{@{}c@{}}ShapeNet\\ Unseen + \\ R-rotate \\R-scale\end{tabular}   & ModelNet  \\\midrule[1pt]
        Graph Interpolation                                                         &14.45                                                    &17.07                                                      &18.23                                                                    &17.07                                                                        &18.39                                                                        &18.36                                                                                &72.51      \\
        \hline
        Graph + Gaussian                                                            &8.926                                                    &11.78                                                      &12.39                                                                    &11.82                                                                        &12.72                                                                        &12.73                                                                                &41.15      \\
        \hline
        FC-based                                                                    & -                                                       & -                                                         &\bf{6.365}                                                               & -                                                                           & -                                                                           & -                                                                                   & -         \\
        \hline
        MSPCG w/o rotate\&scale w/ augmented data                                &6.354                                                    &7.281                                                      &7.250                                                                    &7.263                                                                        &8.037                                                                        &7.993                                                                                &49.30      \\
        \hline
        MSPCG w/o scale                                                             &5.790                                                    &6.854                                                      &6.850                                                                    &6.850                                                                        &7.563                                                                        &7.574                                                                                &47.73      \\
        \hline
        MSPCG w/o rotate                                                            &\bf{4.818}                                               &\bf{6.180}                                                 &6.443                                                                    &7.996                                                                        &\bf{6.618}                                                                   &8.719                                                                                &35.31      \\
        \hline
        MSPCG w/o wCD                                                               &5.201                                                    &6.322                                                      &6.582                                                                    &6.315                                                                        &6.771                                                                        &6.795                                                                                &27.10      \\
        \hline
        MSPCG                                                                       &5.177                                                    &6.323                                                      &6.571                                                                    &\bf{6.312}                                                                   &6.757                                                                        &\bf{6.781}                                                                           &\bf{27.06} \\
        \bottomrule[2pt]
    \end{tabular}
    \caption{Average reconstruction performance in Chamfer Distance, multiplied by $10^4$. Note that our MSPCG achieves the best results in the last two columns, which have the most challenging settings.}
    \label{table:Ablation}
\end{table*}

\begin{figure*}[htb]
    \centering
    \includegraphics[width=\textwidth]{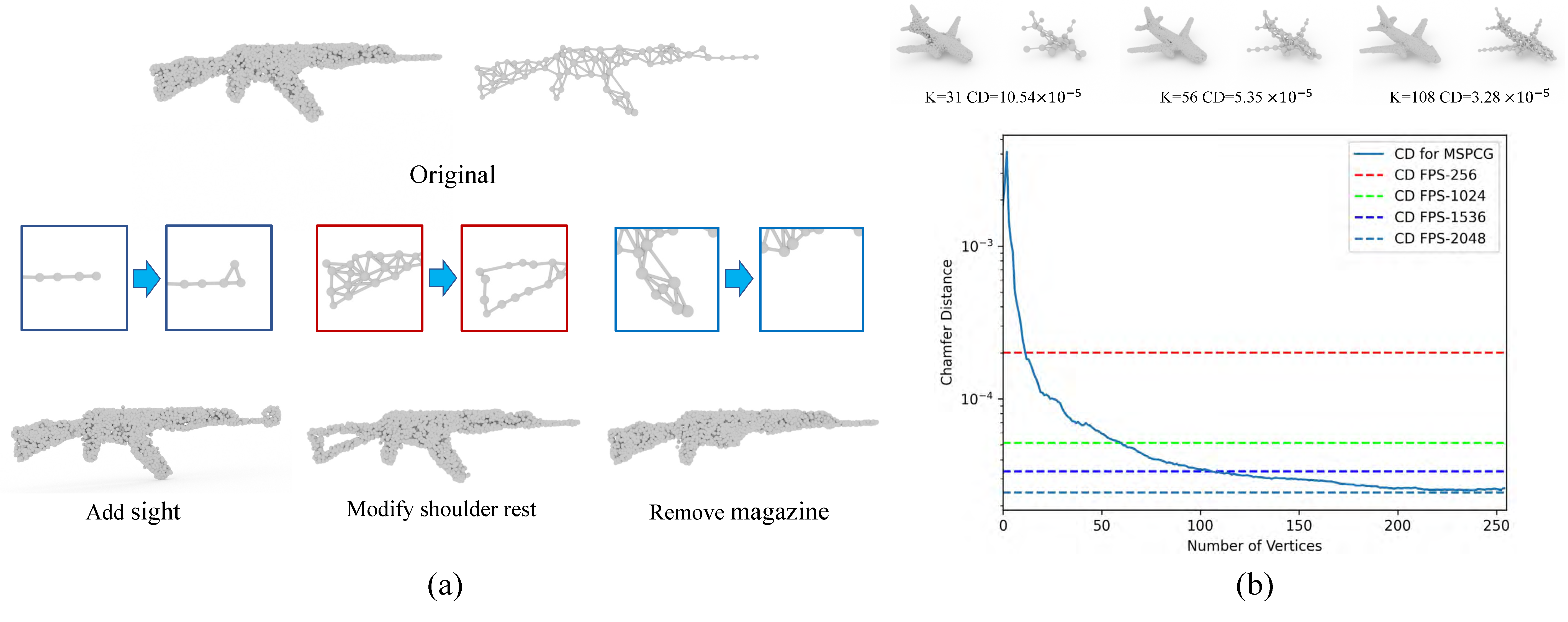}
    \caption{Examples of multiscale editing.}
    \label{fig:Edit}
\end{figure*}

\begin{figure*}[htb]
    \centering
    \includegraphics[width=\textwidth]{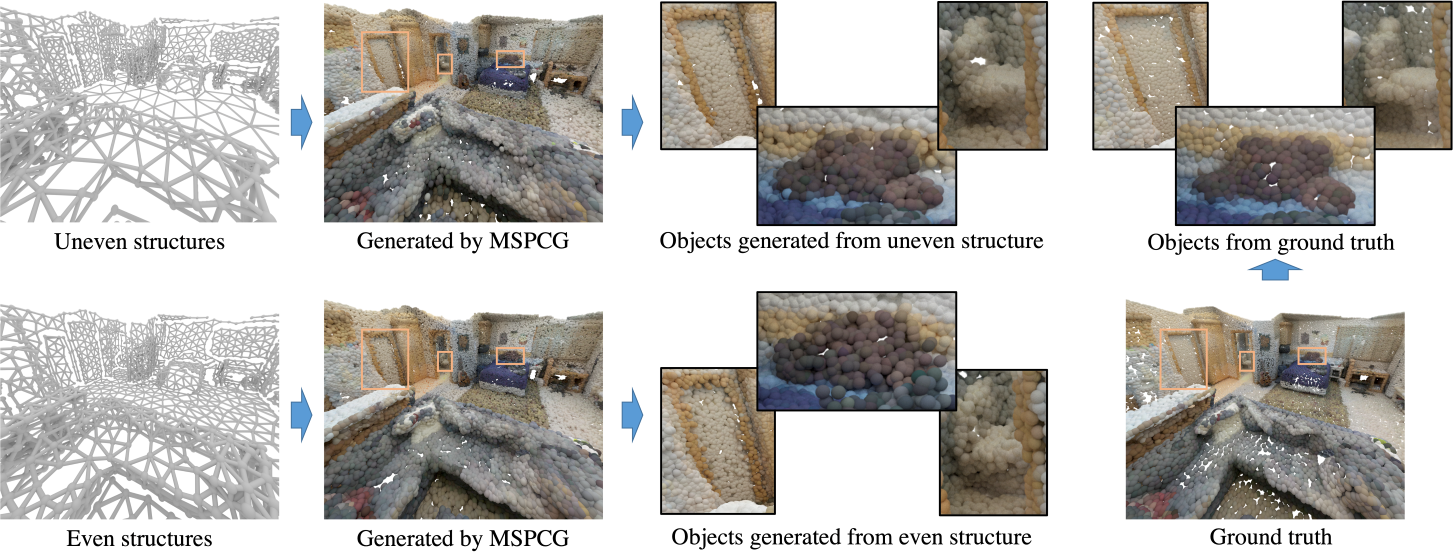}
    \caption{Generation performances of MSPCG on ScanNet dataset. The colors are copied from the closest points in ground truth.}
    \label{fig:SceneGen}
\end{figure*}

To investigate the influences of each part in MSPCG, ablation studies have been down in Tab.~\ref{table:Ablation}. Graph Interpolation is a baseline method that evenly samples points on the edges of MSG. Graph + Gaussian method samples a Gaussian distribution around each vertex in MSG with the variance of its scale factor. The FC-based method uses the decoder of CPCGAN\cite{Yang_Wu_Zhang_Jin_2021}, which transforms each vertex's feature into 64 offsets. As the decoder of CPCGAN cannot generate points with different numbers for different vertices, the FC-based method can only take the MSG with 32 vertices as input.
To get an overall view of the advantages brought by the proposed framework, augmented data are used to train an MSPCG without any rotation or scale-invariant module.

From Tab.~\ref{table:Ablation}, we can find that:
\begin{enumerate}
    \item The learning-based methods have significantly better performance.
    \item The FC-based method exhibits slightly better performance than MSPCG in the 32 vertices' test set. This advantage can be attributed to the fact that the FC-based method focuses solely on one scale during both the training and testing phases. Despite this, MSPCG demonstrates competitive performance in the 32 vertices' test set and boasts superior generalization abilities.
    \item Adopting rotation-invariant and scale-invariant mechanisms is much more efficient than training on augmented data
    \item Removing scale factors causes a performance decline even when objects are not scaled. 
    \item Without the rotation-invariant module, MSPCG performs better in non-rotated cases, indicating that the rotation-invariant mechanism in our MSPCG is not yet perfect. Rotating edges to a canonical pose can lead to significant information loss. However, the full MSPCG exhibits considerable improvement when objects are rotated and maintain competitive performance in non-rotated cases. Therefore, for generalization purposes, we continue to include the rotation-invariant mechanism despite its challenges in certain scenarios.
    \item The weighted Chamfer Distance will slightly help the performance.
\end{enumerate}


\subsection{Multiscale Editing}
The MSPCG method enables control over generation at multiple scales by adding or deleting vertices with different capacities within MSG. Figure \ref{fig:Edit}(a) illustrates some examples of multiscale editing. For instance, adding a single vertex above the barrel creates a detail on a relatively small scale, while deleting a group of vertices modifies the point cloud on a larger scale, such as removing the magazine.

In Figure \ref{fig:Edit}(b), we observe how the reconstruction Chamfer Distance changes with an increasing number of vertices on another object. As more vertices are added to the graph, the reconstruction Chamfer Distance gradually decreases, bringing the generated shape closer to the ground truth. Additionally, the Chamfer Distances between the ground truth and $N$ points sampled from the ground truth using the FPS (Farthest Point Sampling) method are depicted as $\mathit{CD}\ \mathit{FPS-N}$ in Figure \ref{fig:Edit}(b). Notably, with fewer vertices than the sampled points, the point cloud generated by MSPCG can retain more information compared to down-sampled point clouds.

\subsection{Scene Generation}
The MSPCG method can be effectively adapted to larger structured graphs, such as MSGs used for indoor scenes. Figure \ref{fig:SceneGen} illustrates a scene generated by MSPCG. It's worth noting that the model employed in this experiment is the same one trained on ShapeNet as mentioned earlier. 

To highlight the differences between even and uneven structures within each instance, structures are extracted accordingly. For uneven structure extraction, each instance produces an MSG with several vertices equal to 1/16 of the total number of points. Specifically, if an instance has more than 1024 points, it will generate an MSG with only 64 vertices. This approach represents objects with more points more sparsely. 

On the other hand, for even structure extraction, a consistent down-sampling ratio is applied uniformly across all instances. This ratio is carefully selected to ensure that the resulting even structure contains a comparable number of vertices to the uneven structure, approximately 2400 vertices in both cases. Consequently, the point clouds generated from these structures collectively contain approximately 210,000 points.

The results demonstrate that details generated from the uneven structure representation are sharper, cleaner, and closer to the ground truth. This suggests that the proposed MSG could serve as a promising intermediate representation for scene generation in future applications.
    
\section{Limitations}
\label{sec:scope and limitation}
There may be some possible limitations in this study.
The rotation-invariant mechanism is not perfect enough and will reduce the generation qualities when objects are not rotated. Besides, the weighted Chamfer Distance cannot balance all the capacities, which only helps the MSPCG to a small extent. The long-distance information integration is a major limitation in this study that may be addressed in future research. Adding small-scale structures to a pattern shall help the other similar patterns in the same graph to generate similar details ideally, e.g. edit one wing of the airplane to help another wing to generate a similar shape even if no change of structure is adopted on it. Apart from that, the generation of more information modalities such as colors, semantic labels, and normal vectors can be researched in the future.
    
\section{Conclusion}
\label{sec:conclusion}
In this paper, we focused on controlling the generation of 3D details. A Multiscale Structure Graph(MSG) is proposed to represent multiscale structures continuously, and a Multiscale Structure-based Point Cloud Generator(MSPCG) is presented to generate high-quality point clouds from MSG. Various experiments demonstrate that the special designs of MSG and MSPCG help the proposed methods to jointly learn the structures in different scales, perform competitively in point cloud reconstruction, and achieve great generalization ability.


\bibliographystyle{ACM-Reference-Format}
\bibliography{sample-base}

\end{document}